\documentclass[runningheads]{llncs}

\usepackage[T1]{fontenc}
\usepackage{graphicx}
\usepackage{amsmath}
\usepackage{amssymb}
\usepackage{algorithm}
\usepackage{algpseudocode}
\usepackage{xcolor}
\usepackage{orcidlink}
\usepackage[utf8]{inputenc}

\begin{document}

\title{Evolving Explanations: Complexity-Adaptive Refinement of Faithful Explanations for Symbolic Regression}
\titlerunning{Evolving Explanations}

\author{Elisabeth Mayrhuber\inst{1,2}\orcidlink{0009-0006-0471-9997} \and
Ting Hu\inst{3}\orcidlink{0000-0001-6382-0602} \and
Stephan Winkler\inst{1}\orcidlink{0000-0002-5196-4294}}
\authorrunning{Mayrhuber et al.}
\institute{Bioinformatics Research Group, University of Applied Sciences Upper Austria
\email{elisabeth.mayrhuber@fh-hagenberg.at}\and
Computer Vision Lab, TU Wien, Vienna, Austria\and Machine Intelligence and Biocomputing Laboratory, Queen's University, Kingston, Canada}

\maketitle           

\begin{abstract}
Natural-language explanations generated by large language models (LLMs) for machine learning models often seem fluent but are not necessarily faithful to the underlying model. 
We formalize this mismatch as a \emph{faithfulness gap} and propose a complexity-adaptive refinement framework for symbolic regression that iteratively generates and validates explanations until numerical agreement with the target expression is achieved. 
The approach follows a two-stage pipeline: an explainer LLM produces textual descriptions at increasing levels of detail, while an independent validator LLM extracts a symbolic expression that is compared against the original model. 
This process is framed as a discrete search over explanation space, where complexity is minimized subject to a faithfulness constraint.
We evaluate the method on real-world and synthetic datasets, with a total of 250 runs across varying model complexities. 
The explanation refinement loop achieves an overall pass rate of 84.0\%, with most explanations converging at low complexity levels and only a small fraction requiring detailed descriptions.
Our results demonstrate that faithful explanations can be obtained with minimal necessary complexity by explicitly closing the explanation loop through symbolic validation.
\end{abstract}

\section{Introduction}

Large Language Models (LLMs) are widely used to generate text for various tasks, including explanations of machine learning (ML) models\,\cite{zytek_llms_2024,bello_three-level_2025}. 
Despite the widespread use of LLMs for explanation generation, their faithfulness to the underlying model is rarely verified. 
An unfaithful explanation is worse than no explanation: it can cause a domain expert to trust or reject a model for the wrong reasons, violating the core purpose of explainability\,\cite{rudin_stop_2019}.
This is particularly problematic for symbolic regression (SR) models.
Unlike black-box models, SR produces an explicit mathematical expression that \emph{can} in principle be explained precisely, yet LLMs tend to paraphrase, round coefficients, or omit interaction terms, producing explanations that sound plausible but do not faithfully reflect the discovered formula\,\cite{aldeia_interpretability_2022}.

We refer to this as the \emph{faithfulness gap} (Equation\,\ref{eq}): the difference between a model $f$ and the behavior implied by a textual explanation $g$.

\begin{equation}
\label{eq}
\delta(g, f) = \max_{\mathbf{x} \in \mathcal{X}} \lvert g(\mathbf{x}) - f(\mathbf{x}) \rvert
\end{equation}

An explanation is considered faithful if $\delta(g, f) \leq \tau$ for a predefined tolerance $\tau$.
The complexity of a faithful explanation is closely related to the complexity of the underlying symbolic model.
We therefore propose an iterative generation strategy that incrementally increases explanation complexity until $\delta(g, f) \leq \tau$ is met. 
This provides the possibility to find the explanation with the minimum complexity needed to faithfully explain the model.

This iterative explanation generation can be viewed from an evolutionary computation perspective as a search problem in a discrete explanation space. With increasing explanation complexity interpretability is traded for accuracy a phenomena that was already studied in the past\,\cite{aldeia_interpretability_2022}.

We propose a complexity-adaptive explanation refinement loop that iteratively generates explanations from minimal to increasingly detailed complexity levels. The two staged approach works using two functionally independent LLM agents — differentiated by system prompt, temperature, and output format, where one
generates global explanations (explainer) for a given symbolic expression and the second one verifies (verifier) the explanation by extracting a symbolic expression from a given explanation. 
The faithfulness of the explanation can therefore be assessed by calculating the numerical equivalence of the two symbolic expressions.

\subsection{Symbolic Regression and Parsimony}
SR aims to discover an explicit mathematical expression that best fits a given set of observed input-output data points\,\cite{koza_genetic_1994}.
Unlike closed-box regression models, SR produces interpretable expressions that consist of mathematical operators\,\cite{makke_symbolic_2024}. 
Parsimony is an important principle in the context of SR, which describes that among equally accurate models, simpler expressions are preferred\,\cite{burlacu_parsimony_2019}. However, even relatively small symbolic expressions can be hard to interpret for domain experts without a strong mathematical background; therefore, additional explanation methods are required.

\subsection{LLM-Based Explanation of Mathematical Models}
Recent work has explored using large language models to explain mathematical models in natural language\,\cite{zytek_llms_2024,bello_three-level_2025}. 
While such explanations often seem fluent and intuitive, they are not guaranteed to be faithful to the exact mathematical structure of the underlying model or the interpretation of the LLM\,\cite{turpin_language_2023}, especially for nonlinear or interacting terms. 
This motivated an independent verification mechanism that translates textual explanations back into a symbolic expressions to make it comparable to the initial model.

\subsection{Problem Formalisation}
Let $f^*$ denote a symbolic expression discovered from a set of observed input-output data points using SR\,\cite{cranmer_interpretable_2023}. Our goal is to find a natural-language explanation $g$ such that:
\begin{enumerate}
    \item $g$ is understandable to a human (such as a domain expert or an end user),
    \item $g$ corresponds to a symbolic expression $\hat{f}$ extractable from text,
    \item $\hat{f}$ is numerically equivalent to $f^*$ within tolerance $\tau$,
    \item the complexity of $g$ is minimized.
\end{enumerate}

\section{Methodology}
Explanation generation can be seen as a search problem over a space of textual description, where every explanation represents one candidate solution. 
Fitness is determined by its faithfulness regarding the symbolic expression of a SR model. 
Finding the explanation with the minimum required degree of complexity can be formalized as (1+1) evolution strategy\,\cite{back_evolutionary_1996}.

\subsection{Explanation Complexity Levels}
We define a discrete set of explanation complexity levels $C = \{c_0, \dots, c_{\max}\}$, where each level strictly refines the previous one. 
Discrete levels are necessary because natural-language explanations cannot be varied continuously: a prompt must specify a concrete, actionable instruction for the LLM to follow, and marginal increases in verbosity do not reliably translate to proportional increases in mathematical precision.
Discrete steps also make the (1+1)-ES formulation tractable, providing a well-defined step operator $c \mapsto c+1$ and a finite search space.
To generate explanations of different complexity levels, the prompt that is sent to the LLM is adjusted accordingly. 
We use five levels ($c_0$--$c_4$) (Table\,\ref{tab:complexity}) as a practical granularity that spans the full range from a single trend sentence to a complete term-by-term analysis, while keeping the worst-case number of LLM calls per instance bounded at ten (two validator attempts per level).
Example\,\ref{ex:prompt} shows the complexity indication in a explanation generation prompt for complexity level $c_2$.

\begin{table}[ht]
\centering
\begin{tabular}{ll}
\hline
Level & Description \\
\hline
$c_0$ & Overall trend and direction of effects \\
$c_1$ & Individual terms and directional influence \\
$c_2$ & Relative magnitude of effects \\
$c_3$ & Nonlinearities and interactions \\
$c_4$ & Full term-by-term mathematical interpretation \\
\hline
\end{tabular}
\caption{Hierarchy of explanation complexity levels defining progressively richer semantic descriptions of a symbolic expressions.}
\label{tab:complexity}
\end{table}

\begin{example}
\label{ex:prompt}
    "... Name each term, state its direction, AND describe its relative, magnitude and importance compared to other terms. Mention whether effects are strong, moderate, or weak. ..."
\end{example}

\subsection{The Refinement Loop}
\begin{figure}[ht]
    \centering
    \includegraphics[width=\linewidth]{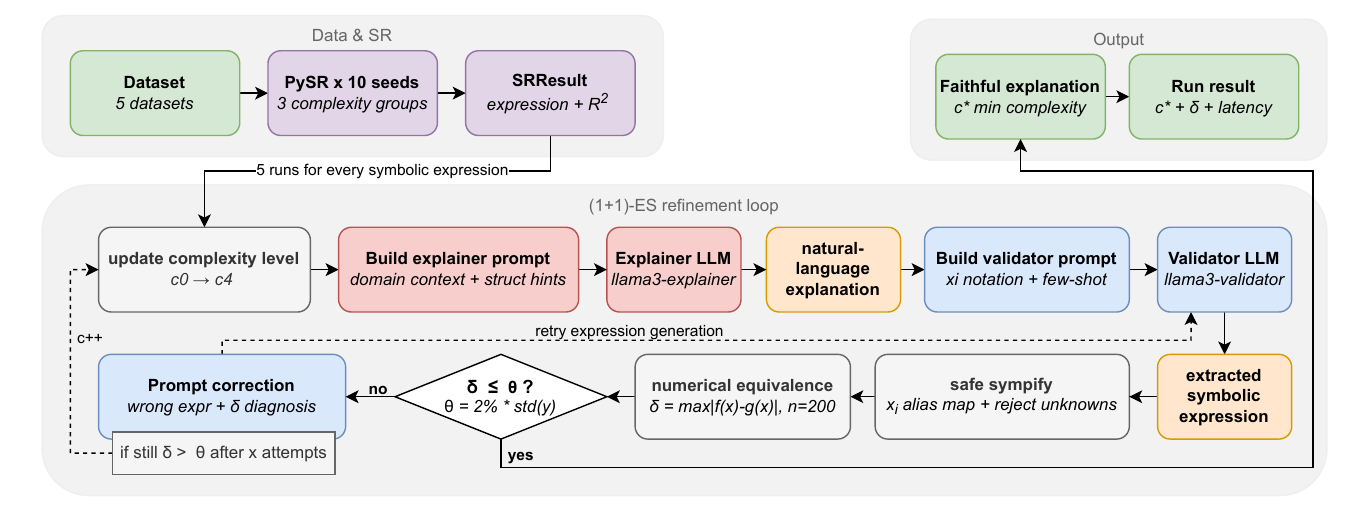}
    \caption{Proposed workflow for evolutionary explanation generation.}
    \label{fig:workflow}
\end{figure}

Our approach iteratively generates explanations of increasing complexity levels (Table\,\ref{tab:complexity}). 
Each explanation is evaluated regarding its faithfulness by comparing the results of 200 uniformly distributed feature inputs for the original model and the extracted symbolic expression from the evolved explanation. 
In case the extraction of a symbolic expression from the explanation fails the process is repeated $R$ times until the complexity is increased (Figure\,\ref{fig:workflow}). 
If the faithfulness of the explanation on the highest complexity degree is not sufficient the explanation generation fails (Algorithm\,\ref{alg:workflow}). 
To ensure robustness, extracted expressions are parsed using a restricted SymPy\,\cite{meurer_sympy_2017} environment, only allowing a certain set of operators.

\begin{algorithm}
\caption{Complexity-Adaptive Explanation Refinement}
\label{alg:workflow}
\begin{algorithmic}[1]
\Require Symbolic expression $f^*$, complexity levels $C = [c_0 \ldots c_{\max}]$,
         tolerance $\tau$, feature bounds $\mathbf{b}$, 
         max complexity steps $T$, extraction retries $R$
\Ensure Faithful explanation $g^*$ at minimal complexity, or failure

\For{$t = 0, 1, \ldots, \min(|C|-1, T)$}
  \State $c \leftarrow C[t]$
  \State $g \leftarrow \textsc{Explain}(f^*, c)$
    \Comment{Explainer LLM at complexity $c$}

  \State $\hat{f} \leftarrow \varnothing$
  \State $r \leftarrow 0$

  \Repeat
    \State $\hat{f} \leftarrow \textsc{Extract}(g)$
      \Comment{Validator LLM: text $\rightarrow$ SymPy expression}
    \State $r \leftarrow r + 1$
  \Until{$\hat{f} \neq \varnothing \;\lor\; r = R$}

    \State $\delta \leftarrow 
      \max_{\mathbf{x} \in \mathbf{X}}
      |\hat{f}(\mathbf{x}) - f^*(\mathbf{x})|$
      \Comment{Validation on $200$ samples}

    \If{$\delta \leq \tau$}
      \State \Return $(g, c, \delta)$
      \Comment{Faithful explanation found at level $c$}
    \EndIf
\EndFor

\State \Return \textbf{failure}
\end{algorithmic}
\end{algorithm}

\subsection{EC Formalisation}
We formalize the refinement loop as a discrete $(1+1)$ evolution strategy. 
Each explanation attempt corresponds to an individual, and the complexity level serves as the mutation step size. 
Fitness is defined as the faithfulness $\delta$, which is the maximum numerical discrepancy between the candidate and target expressions evaluated on 200 samples. 
An individual is only accepted if its faithfulness is lower than the tolerance $\tau$.

\subsection{Experimental Setup}
We evaluate our method on a set of real-world and synthetic regression datasets. For each dataset, a symbolic regression model is fitted using PySR\,\cite{cranmer_interpretable_2023}. 
The datasets are selected to create SR models of increasing structural and mathematical difficulty, enabling a controlled analysis of how explanation faithfulness depends on explanation detail rather than dataset-specific artefacts (Table\,\ref{tab:datasets}). 
The refinement loop (Algorithm\,\ref{alg:workflow}) is executed with ten independent PySR fits, in three different model complexity stages per dataset and five LLM runs per fit (50 runs per dataset, 250 total). 
All experiments use adaptive tolerance $\tau = 0.02 \cdot \sigma_y$ and two validator attempts per complexity level.

All experiments were executed on commodity hardware (13th Gen Intel(R) Core(TM) i5-13600H (2.80 GHz), 64 GB RAM) using Ollama as the LLM serving stack\,\cite{grattafiori_llama_2024}. 
Both agents are based on the same base model, namely Meta Llama~3 8B Instruct, served in two separate Docker containers. 
The explainer container uses a Modelfile with a natural-language system prompt and temperature $T=0.3$; the validator container uses a structured JSON output prompt with $T=0.1$. 
No fine-tuning was applied to either model.

\begin{table}[ht]
\centering
\begin{tabular}{lp{2.2cm}p{2.2cm}p{4.2cm}}
\hline
Dataset & Type & Complexity & Motivation / Reference \\
\hline
Diabetes &
Real-world &
Low &
Linear or near-linear structure; early convergence \cite{efron_least_2004} \\

Friedman \#1 &
Synthetic &
Medium &
Standard nonlinear benchmark with interactions \cite{friedman_multivariate_1991} \\

Friedman \#2 &
Synthetic &
High &
Strong nonlinearities; stress test for explanations \cite{friedman_multivariate_1991} \\

Synthetic Polynomial &
Synthetic &
Medium &
Known ground truth enables faithfulness verification \cite{koza_genetic_1994} \\

Synthetic Trigonometric &
Synthetic &
High &
Interaction-heavy trigonometric structure \cite{koza_genetic_1994} \\
\hline
\end{tabular}
\caption{Overview of datasets used for evaluation.}
\label{tab:datasets}
\end{table}

\section{Results}
The refinement loop achieves an overall pass rate of $84.0\%$ across all 250 runs (Table~\ref{tab:results}).

\begin{table}[ht]
\centering
\begin{tabular}{p{3.8cm}p{1.8cm}p{1.4cm}p{1.4cm}cc}
\hline
Dataset & \centering SR $R^2$ & \centering Nodes & \centering Pass (\%) & Median $c^*$ & Runtime (s) \\
\hline
Diabetes (linear)       & \centering 0.40--0.45 & \centering 9--13  & \centering 80.0  & 1          & 311 \\
Friedman \#1            & \centering 0.70--0.79 & \centering 12--17 & \centering 62.0  & 2          & 529 \\
Friedman \#2            & \centering 0.997      & \centering 3      & \centering 100.0 & 1          & 143 \\
Synthetic polynomial    & \centering 0.990      & \centering 7      & \centering 88.0  & 2          & 332 \\
Synthetic trigonometric & \centering 0.12--0.30 & \centering 5--17  & \centering 90.0  & 1          & 267 \\
\hline
\textbf{Overall}        & \centering --         & \centering --     & \centering 84.0  & \textbf{1} & \textbf{316} \\
\hline
\end{tabular}
\caption{Benchmark results for $n=50$ runs per dataset (10 SR fits $\times$ 5 LLM runs). $c^*$: minimum explanation complexity achieving $\delta \leq \tau$. Pass: fraction of runs converging within $c \leq 4$. Ranges in SR\,$R^2$ and nodes reflect variation across the 10 independent PySR fits; single values indicate all fits converged to the same expression. The runtime reflects the average runtime per dataset on commodity hardware.}
\label{tab:results}
\end{table}

\begin{figure}[ht]
    \centering
    \includegraphics[width=\linewidth]{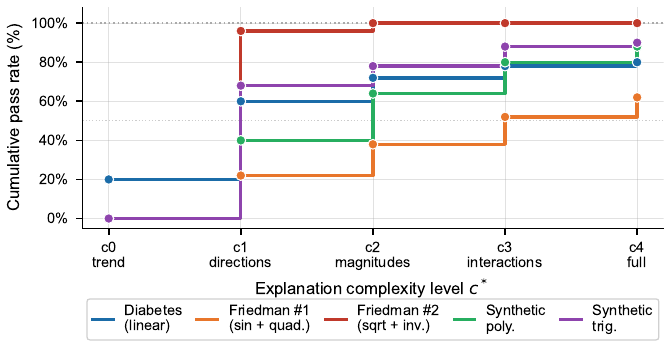}
    \caption{Cumulative pass rate for every data set and complexity level, showing that Friedman \#2 converging at $c^2$ latest and Friedman\#1 having the smallest cumulative pass rate.}
    \label{fig:fitness_curves}
\end{figure}

\begin{figure}[ht]
    \centering
    \includegraphics[width=\linewidth]{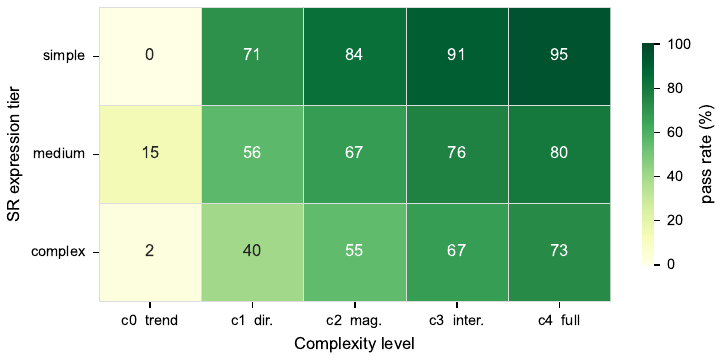}
    \caption{Pass rate (\%) for each combination of SR expression complexity tier (rows) and explanation complexity level (columns). Darker cells indicate that more faithful explanations could be generated. Simple expressions achieve high pass rates already at $c_1$, while complex expressions require higher complexity levels before convergence, confirming that $c^*$ scales with SR model complexity.}
    \label{fig:sr_exp_complexity}
\end{figure}

\begin{figure}[ht]
    \centering
    \includegraphics[width=\linewidth]{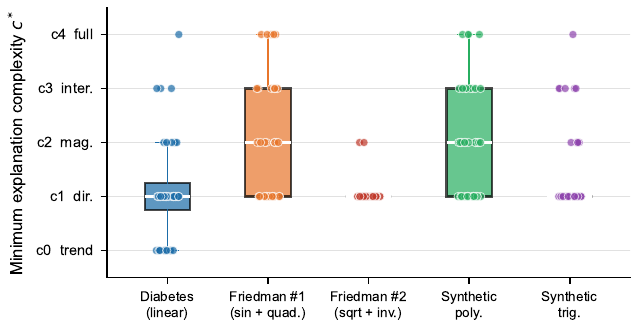}
    \caption{Distribution of the minimum complexity level needed for each test run.}
    \label{fig:cstar_distribution}
\end{figure}

The majority of runs converge at low complexity: $57.2\%$ already succeed at or before complexity level $c_1$, and $70.4\%$ at or before $c_2$. 
Only $16.2\%$ of converged runs require $c^* \geq 3$. 
This indicates that the system generally favors simple, interpretable explanations and only increases complexity when explanations at low complexity levels don't succeed (Fig.~\ref{fig:fitness_curves}).
The distribution of $c^*$ is heavily concentrated at $c_1$, which accounts for 63.3\% of all converged runs, followed by $c_2$ (15.7\%), $c_3$ (11.0\%), and $c_0$ (4.8\%).
This reflects the structure of the complexity ladder: $c_0$ (a single trend sentence) rarely provides enough mathematical detail for reliable extraction, while $c_1$ provides a good balance between simplicity and extractability for the range of expressions PySR produces on these datasets (Fig.~\ref{fig:cstar_distribution}).

\subsubsection{Diabetes:} PySR recovers linear expressions of complexity 9--13 with $R^2 \in [0.40, 0.45]$, consistent with the known near-linear structure of this dataset. 
The pass rate for the explanation generation is 80.0\% with median complexity level $c^* = 1$.
 
\subsubsection{Friedman \#1:} With $R^2 \in [0.70, 0.79]$, PySR partially recovers the sinusoidal interaction term but struggles with the exact $\sin(\pi x_0 x_1)$ structure due to operator interactions. 
The pass rate for explanation generation is 62.0\%, which is the lowest across all datasets with a median complexity level $c^* = 2$, indicating that richer explanations are required before the validator can reconstruct even an approximate expression.
 
\subsubsection{Friedman \#2:} Runs on this dataset achieve a pass rate of 100\% with median explanation complexity level $c^* = 1$.
PySR consistently discovers the expression $x_1 \cdot x_2$, which represents a simplified but highly faithful three-node expression ($R^2 = 0.997$), rather than the full ground-truth $\sqrt{x_0^2 + (x_1 x_2 - 1/(x_1 x_3))^2}$. 
This compact expression is trivially explained at $c_1$ and extracted without any errors.
 
\subsubsection{Synthetic polynomial:} The pass rate of 88.0\% with a median $c^* = 2$ aligns with the expected difficulty of a expression with a cross-product interaction term. 
PySR consistently finds a correct expression ($R^2 = 0.990$), and $c_2$  is needed for the validator to preserve the interaction of coefficients accurately.
 
\subsubsection{Synthetic trigonometric:} Despite a low $R^2 \in [0.12, 0.30]$ the pass rate reaches 90.0\% with median $c^* = 1$. This demonstrates a key property of the refinement loop: \emph{faithfulness is independent of model quality}. The loop validates whether the explanation faithfully captures whatever expression PySR found, regardless of whether that expression is a good model of the data.

\subsection{SR Expression Complexity versus Explanation Complexity}
\begin{figure}[ht]
    \centering
    \includegraphics[width=\linewidth]{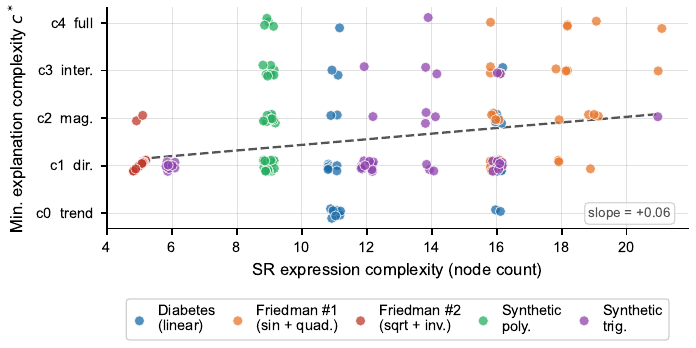}
    \caption{Minimum needed explanation for different SR expression complexity, showing only a small direct relationship between model and explanation complexity.}
    \label{fig:complexity_vs_cstar}
\end{figure}

Across all datasets, SR expression node count correlates positively with $c^*$ (Pearson $r = 0.28$, $p < 0.01$, Fig.~\ref{fig:complexity_vs_cstar}). 
The relationship is strongest within Friedman \#1 ($r = 0.43$) and synthetic trigonometric ($r = 0.35$), where PySR finds expressions of varying complexity across fits. 
Datasets where PySR consistently recovers a single expression (Friedman \#2: always a node count of 3; synthetic polynomial: always a node count of 7) show no within-dataset variation. 

\subsection{Variance Decomposition}
\begin{figure}[ht]
    \centering
    \includegraphics[width=\linewidth]{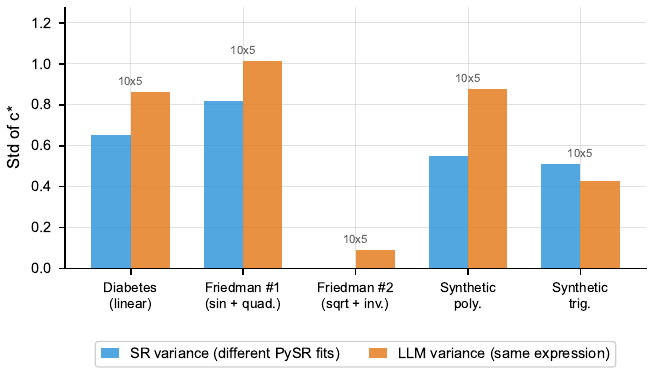}
    \caption{SR variance (blue) versus LLM variance (orange) showing that the LLM is the bottleneck in 4 out of 5 cases.}
    \label{fig:variance_decomp}
\end{figure}

Fig.~\ref{fig:variance_decomp} decomposes the variance of $c^*$ into SR-fit variance (does $c^*$ differ across different PySR runs?) and LLM variance (does $c^*$ differ across repeated explanation attempts for the same expression?). 
LLM variance dominates in four of five datasets, with SR variance only exceeding LLM variance for the synthetic trigonometric dataset. 
This suggests that the primary bottleneck for faithful explanation is \emph{LLM extraction stochasticity} rather than PySR instability.

\subsection{Runtime}
\begin{figure}[h!]
    \centering
    \includegraphics[width=\linewidth]{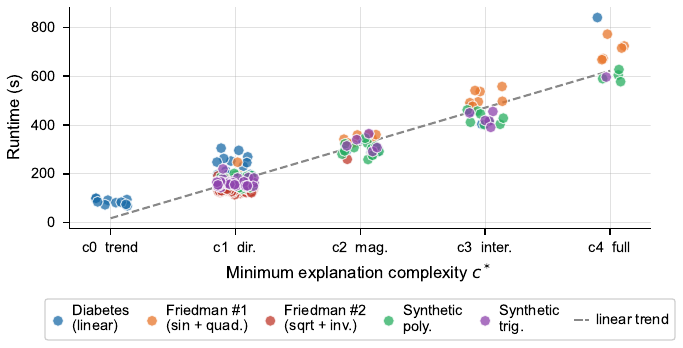}
    \caption{Runtime for all complexity levels and datasets on commodity hardware.}
    \label{fig:latency}
\end{figure}

Mean per-run runtime is 316.4\,s overall, ranging from 68.53\,s (converges immediately) to 858.59\,s (requires multiple complexity escalations).
The linear relationship between $c^*$ and runtime is visible in Fig.~\ref{fig:latency}: each complexity escalation adds 60--105\,s depending on the dataset, reflecting the combined cost of re-prompting the explainer and re-running the validator. 
The experiments were executed on commodity hardware.

\begin{figure}[h!]
    \centering
    \includegraphics[width=\linewidth]{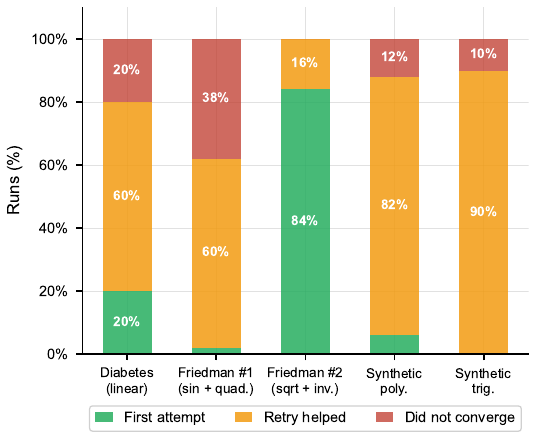}
    \caption{Success rates of faithful LLM explanation generation where green show the runs that converged at the first validation try, yellow the explanations that needed more than one validation attempt and red did not converge at all.}
    \label{fig:retry_success}
\end{figure}

\subsection{Retry Effectiveness}
Fig.~\ref{fig:retry_success} shows how effective re-running the validator is for each dataset. 
The majority converge on the first validator attempt. 
The correction prompt (second attempt with the wrong expression and a diagnosis of the error magnitude) improves pass rate, particularly for datasets where the validator tends to produce structurally correct but numerically imprecise extractions (diabetes, synthetic polynomial).
The residual 16.0\% non-convergence rate reflects cases where the expression is either too complex for the LLM to describe faithfully at any level, or where PySR found a degenerate expression that cannot be explained without reproducing it verbatim.

\section{Discussion}
The idea that more complex SR expressions require richer natural-language explanations to achieve faithful extraction is supported by the presented results.
We observe a positive correlation between SR node count and $c^*$ ($r = 0.28$), even within the relatively narrow complexity range of this benchmark (3–17 nodes) and despite cases where PySR produces identical expressions across runs.

\subsection{Faithfulness Decouples from Model Quality}
The synthetic trigonometric dataset provides the clearest demonstration that faithfulness is a property of the explanation--expression pair, not of the expression--data pair.
With $R^2$ as low as 0.12, PySR's expressions are poor models of the underlying function, yet 90\% of explanation runs converge.
This decoupling is a desirable property: users receive an accurate description of what the model actually computes, regardless of whether the model is scientifically valid.

\subsection{Friedman \#2: Parsimony as a Double-Edged Sword}
The consistent discovery of $x_1 \cdot x_2$ for Friedman \#2 illustrates a tension inherent in parsimony-penalised SR.
The expression is mathematically faithful to what was fitted and trivially explainable at $c_1$.
However, it discards the structural information in the ground truth ($\sqrt{x_0^2 + (x_1 x_2 - 1/(x_1 x_3))^2}$) that a domain expert would find meaningful.
High faithfulness of a simple but wrong expression is not the same as a faithful explanation of the true model. 

\subsection{LLM Variance as the Dominant Bottleneck}
The variance decomposition reveals that LLM extraction contributes more to variation in $c^*$ than differences in SR fitting for four out of five datasets.
This has a clear practical implication: improving the validator prompts (e.g. more precise feature mapping instructions, richer few-shot examples, or fine-tuned models). 
The only exception is the synthetic trigonometric dataset, where PySR produces expressions with genuinely different levels of complexity (5–17 nodes), so variation in SR fitting naturally plays a larger role.

\subsection{Limitations and Future Work}
The current benchmark uses a single LLM pair (llama3-explainer / llama3-validator) without ablation over model size or architecture.
The correlation between SR complexity and $c^*$, while statistically significant, is moderate ($r = 0.28$) and may reflect prompt sensitivity as much as genuine structural complexity.
Future work should include: 
(i) the mixed-complexity experiment with controlled node counts across SR fits to isolate the complexity--faithfulness relationship; 
(ii) ablation over LLM model size and temperature; 
(iii) a human evaluation study to validate whether $c^* = 1$ explanations are genuinely useful to domain experts, or whether enforcing higher complexity produces better-understood explanations at the cost of brevity.

\section{Conclusion}
We introduce a complexity-adaptive framework for generating faithful natural-language explanations for SR models. 
By explicitly closing the explanation loop through symbolic validation, our method ensures that explanations are not only fluent but also numerically consistent with the underlying model. 
The iterative refinement process identifies the minimum level of explanation complexity required to achieve faithfulness, thereby balancing interpretability and precision.
Our results show that, in most cases, relatively simple explanations are sufficient. This supports the idea that faithful explanations do not necessarily require full mathematical detail, but rather the appropriate level of abstraction matched to the model. 
At the same time, the observed variability introduced by the extraction step highlights that LLM-based explanation pipelines benefit significantly from explicit verification mechanisms.
Overall, our work highlights the importance of coupling explanation generation with validation when using LLMs for explainability. 
Ensuring faithfulness through closed-loop verification points towards more trustworthy and interpretable explanation methods for mathematical models.

\begin{credits}
\subsubsection{\ackname} This research was funded by the Austrian Science Fund (FWF) [\url{https://doi.org/10.55776/DFH34}].
\end{credits}

\subsubsection{\discintname}
The authors have no competing interests to declare that are relevant to the content of this article.

\bibliographystyle{splncs04}
\bibliography{references}
\end{document}